# ANALYSIS OF THE TRANSPORT MODEL THAT APPROXIMATES DECISION TAKER'S PREFERENCES

*V.Ya. Vilisov*
*University of Technology, Russia, Moscow Region, Korolev*
*vvib@yandex.ru*

**Abstract.** Paper provides a method for solving the reverse Monge-Kantorovich transport problem (TP). It allows to accumulate positive decision-taking experience made by decision-taker in situations that can be presented in the form of TP. The initial data for the solution of the inverse TP is the information on orders, inventories and effective decisions take by decision-taker. The result of solving the inverse TP contains evaluations of the TP's payoff matrix elements. It can be used in new situations to select the solution corresponding to the preferences of the decision-taker. The method allows to gain decision taker's experience, so it can be used by others. The method allows to build the model of decision taker's preferences in a specific application area. The model can be updated regularly to ensure its relevance and adequacy to the decision taker's system of preferences. This model is adaptive to the current preferences of the decision taker.

**Introduction**

Modern information technologies provide new possibilities for managing complicated social and economic objects. However, computers' computational abilities very often outpace technological capabilities of the existing management procedures. Thus, the tendency of "burdening" the computers with a larger number of managerial functions, which was predicted by the fathers of the management science [3, 5, 11], is well preserved and gradually turns into a common practice. Modern management includes different tools such as elements of artificial intellect (expert systems, genetic algorithms, neural nets etc.), simulation modelling tools, adaptive management ideas and so forth.

This work is devoted to some issues of the enterprise adaptive management reviewed on the basis of the transport-related business processes. Ideas of adaptive management related to the social and economic objects are developing for quite a while already [1, 2, 3]. Some authors [1, 9] think that the ideal economics (and its all-level objects) should possess a high degree of adaptability like living creatures in nature. Here it means that social and economic objects should be able to adapt to the unfavorable external disturbances by rebuilding their structure or by changing parameters. At that it is the social component that plays the adaptability role in relation to the environment (it consists of managers, operators etc., i.e. the people that take the decisions - Decision Takers). Currently, the algorithmic component is not yet fully developed, only playing a role of the computational support. Within the framework of this adaptation technology, it is only the decision takers that accumulate the adaptability experience, so that when they are changed or are absent in some form or way, the whole system's experience is lost thus decreasing the effectiveness of its function.

The work considers another adaptation aspect where the adaptability experience stays within the system even when the decision taker is taken away or when he/she is absent temporarily. This approach is considered within the context of the transport system management problems. The experience is kept by the economic and mathematical models, parameters (sometimes it could be even the structure itself), which are set according to the decisions taken by the decision taker in the specific situations. Thus, these models approximate the preferences of the decision taker for the real-time situation by taking into account the uncertainty and instability of the environment. It is possible to say that they preserve the experience of decision taker. Still, like every preservation, the experience of decision taker has a limited validity period due to the instability of the internal characteristics of the system and the environment. Such preservations can be used by the system either without direct participation of the experience donor (the decision taker), with his/her minimal participation or with the participation of other decision takers that manage the same objects. Such accumulators of the decision taker's positive experience show his/her preferences as criteria and objective functions. Apart from the property that separates the experience (objective preferences) of the decision taker from its bearer, the technology under review plays another important role being the convolution of many real-time objective indices into the scalar objective function that approximates the objective indices vector. It is necessary to mention that part of these objective indices can be considered by the decision taker only on the subconscious level.

Some scientists had already mentioned the necessity of formal characterization of the decision taker's experience and its further usage in the management procedures [7, 8]. And according to Herbert Simon [11] all situations, involving decision-taking when managing social and economic objects can be divided into the structured and non-structured ones. All procedures should gradually become structured, being performed only by the computations means of the Corporate Information Systems (CIS).

There also exist other approaches for the formal characterization of the decision taker's experience, such as expert systems and neural nets. Still, their practical application technique does not yet provide effective ways for managing the business processes under review.

As a rule, modern enterprise's CIS [11] includes some elements of ERP-, APS- and MES-systems. The algorithms that are considered in this work are directed at developing and amending the functions of APS-systems.

Within the framework of the suggested technology the object management is performed in the two-circuit configuration. In the first circuit we perform the adjustment of the model parameters basing on the decisions taken by the manager (decision taker), while in the second one we perform direct object management basing on the model. In this scheme the first circuit works according to the manager's natural tempo while the second one works in the rhythm of the managing processes. Thus, the high intensity of the data flow that is present during the working management of the processes does not reduce the quality of approved managerial decisions. However, here (according to H. Simon [11]), we see the reduction of the manager's bounded rationality impact upon the management quality.

The work investigates usage of transport models (its peculiarities and properties) as the preservatives of the positive management experience within the transport systems. Materials of this work develop the author's earlier research conducted in the field of the adaptive management within the social and economic and technical systems [4, 13, 14]. The performed analysis is based on the simulation experiment, using the MS Excel's Data Analysis add-in. For the purposes of better result visualization, but without losing the communality, the research deals with the minimum dimensionality models.

**Direct and Reverse Formulation of the Transport Problem**

Transport problem model (TPM) is the special case of the linear programming model [12]. To solve the transport problem means to find the number of goods ($x_{ij}$) that are sent from the point of departure to the destination points. Here the *plan's optimality criterion is the minimum amount of the total transportation cost*. Usually we know such source data as the stock quantity ($a_i$) in the point of departure and the goods demand rate ($b_j$) per each destination point. We also know the commodity unit transportation cost matrix ($c_{ij}$) from $i$-th departure point to $j$-th destination point.

Historically, these linear programming problems were included into a specific group due to their special structure that allows solving them more effectively using specially developed hand-calculation methods. However, today, using the existing software of modern and powerful computers, TPM can be solved as a usual direct linear programming problem (LPP). At that the problem's characteristics will be reflected only in its source setting. Further we shall show how we can represent the TPM source setting as a standard LPP. In this case, for solving the reverse LPP (RLPP) we shall use one of its solution algorithms [4, 12], so that after performing a reverse transition we could obtain the estimates of TPM coefficients.

Usually, when compiling the transportation plan, the total transportation cost is minimized. But in real time the problem not only contains many criteria, but is also flexible in time (unstable). Therefore it is difficult to state a priori whether it is the total cost or the time that is the dominating factor, or maybe it is some other non-formalized indices, about which only the decision taker knows. This is why we think that it would be highly useful to find some common convolution that would include decision taker's integral preferences in relation to the multitude of alternatives.

In comparison with the standard LPP, the TPM's peculiarity lies in the bigger dimensionality even with the fewer number of the departure and destination points, at the same time possessing a higher sparsity of the solution matrix (many zero cells in the solution matrix). Standard form of the transport model is usually used when describing situations involving homogeneous goods, which can be delivered from any $A_i$ point to a random $B_j$ point.

Still, the situations involving heterogeneous goods happen more frequently when dealing with practical transport decisions. Here the stocks and inquiries change frequently and the coefficient matrix $c_{ij}$ is unstable (it depends on season, daily and random fluctuations of the transport network). Moreover, in the real time (where the optimization models are used), the transport problem is usually solved together with optimizing the workload of the existing car park.

Despite the differences between the classic set-up of the transport problem and the big number of real-time situations, there are some cases where we can use TPM. For the purposes of TPM's higher adequacy, let us agree that the transport cost of one commodity unit from a *i*-th point to a *j*-th point shall stand not only for a money equivalent, but for some general expenses that are taken into consideration by the decision taker when choosing the transport plan. In this case the solution of the reverse transport problem shall be the following: basing on the situation monitoring (stocks and inquiries) as well as the transport plans that are defined as positive according to their operational results, we shall build $\hat{c}_{ij}$ estimates, according to which it will be possible to make transport plans for the new situations, thus solving a new transport problem. This will allow to automate the planning process either by completely replacing the decision taker (within the stable environment) or by significantly decreasing the decision taker's workload in the part of the transport plan compilation. However, we shall preserve the compliance of the transport plans with the decision taker's preferences and his/her practical experience. When using the adjusted model for planning within the flexible environment, prior to its sending for execution, the model-type plan variant can be given to the decision taker for approval or corrections.

For the purposes of discussion, the decision taker can be defined as a "black box" that transforms the stocks and inquiries vectors (that define the decision making situations - DMS) into the transport plan (see Picture 1).

Solution of the reverse transport problem (RTP) is the following: by monitoring the situations and actions of the decision taker, we shall discover the decision taker's preference system (see Picture 2). Basing on this we shall solve the direct transport problem using one of the methods.

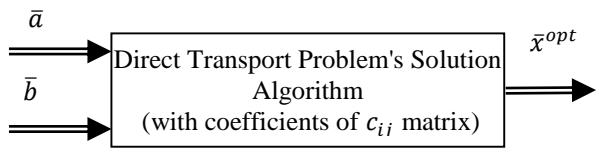

Pic. 1. Solution Module for the Direct TP

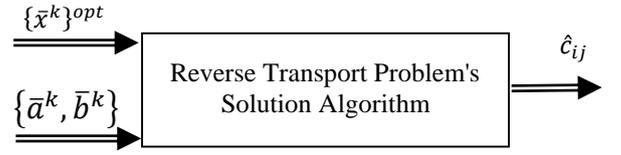

Pic. 2. Solution Module for the Reverse TP

The condition (DMS) is defined by the number of delimitation coefficients $\|a_i, b_j\|_{mn}$:

$$\sum_{j=1}^{n} x_{ij} = a_i, \quad i = \overline{1,m}; \quad (1)$$

$$\sum_{i=1}^{m} x_{ij} = b_i, \quad j = \overline{1,n}; \quad (2)$$

$$x_{ij} \geq 0, \quad i = \overline{1,m}; \quad j = \overline{1,n}. \quad (3)$$

For the TPM, $\bar{x}^k$ solution per each DMS can be represented as the R-values matrix $\bar{x}^k = \|x_{ij}^k\|_{mn}^N$. In this case, the approximation of the decision taker's preferences performed by the transport problem model lies in the evaluation of $c_{ij}$ coefficients, pertaining to the target function:

$$L(\bar{x}) = \sum_{i=1}^{m} \sum_{j=1}^{n} c_{ij} x_{ij}, \quad (4)$$

which can be defined as general transport expenses when transporting a commodity unit from *i*-th departure point to *j*-th destination point. When solving a direct TP, we shall provide $L(\bar{x}) \to \min_{x_{ij}}$.

**Transformation of the Transport Problem into the Linear Programming Problem**

For the purposes of solving the RTP, let us *maximally convert* TPM into the *LPP* using inequality constraints. For this let us transform equality constraints into inequality constraints, minimum OF into the maximum OF, also performing other transformations. These transformations are necessary because the reverse problem solution algorithm exists [4] for the mentioned LPP, not existing for the TPM, which is the specific type of problem.

Given that within the set of constraints of $(m + n)$ equations only $(m + n - 1)$ equations are linearly independent (one equation is redundant, as the sum of equations derived according to the lines of the payment matrix equals to the sum of equations that are derived according to the columns - order-inquiry balancing property), let us express $(m + n - 1)$ of variables (basic ones) via the rest (free ones). For the purposes of certainty, these variables shall be $x_{i1}, x_{1j}; i = \overline{1,m}; j = \overline{2,n}$. Let us express $x_{i1}, x_{1j} (i = \overline{1,m}; j = \overline{2,n})$ variables via the rest:

$$x_{11} = a_1 - \sum_{j=2}^{n} b_j + \sum_{i=2}^{m}\sum_{j=2}^{n} x_{ij}; \quad (5)$$

$$x_{i1} = a_i - \sum_{j=2}^{n} x_{ij}, \quad i = \overline{2,m}; \quad (6)$$

$$x_{1j} = b_j - \sum_{i=2}^{m} x_{ij}, \quad j = \overline{2,n}; \quad (7)$$

Having also transformed the objective function and by placing the values of basic variables into the source OF (4) using free variables (5)-(7), we shall obtain OF and corresponding delimitations represented as *maximum* LPP. Here all OF coefficients should be multiplied by (-1), which, with new OF *maximization* shall correspond to the *minimization of the source OF*.

$$L(\bar{x}) = \sum_{i=2}^{m}\sum_{j=2}^{n} \tilde{c}_{ij} x_{ij} \to \max_{x_{ij}}, \quad (8)$$

where $\tilde{c}_{ij} = -(c_{11} - c_{i1} - c_{1j} + c_{ij})$

$$\sum_{j=2}^{n} b_j - a_1 - \sum_{i=2}^{m}\sum_{j=2}^{n} x_{ij} \leq 0; \quad (9)$$

$$\sum_{j=2}^{n} x_{ij} - a_i \leq 0, \quad i = \overline{2,m}; \quad (10)$$

$$\sum_{i=2}^{m} x_{ij} - b_j \leq 0, \quad j = \overline{2,n}; \quad (11)$$

$$x_{ij} \leq 0; \; i = \overline{2,m}; \; j = \overline{2,n}. \quad (12)$$

$\tilde{c}_{ij}$ coefficients should be estimated according to the solution algorithm of the reverse linear programming problem (RLPP). For the purposes of solving the direct transport problem on the basis of the built model, it is necessary, using the appearing DMS (i.e. $\bar{a}, \bar{b}$ vector population) to solve the linear programming problem (8)-(12). As a result of this, we shall find variables $(m-1) \times (n-1)$, while the rest $(m + (n-1))$ should be calculated according to the formulae (5)-(7).

This research uses the point-like algorithm, which is one of the RLPP solving algorithms. Its essence will be shown below using one of the examples.

**Researching Properties of the Reverse Transport Problem**

For the purposes of simplicity and result interpretation, let us consider a TP possessing two points of departure (*m*=2) and three points of destination (*n*=3). All data for this problem is provided in Table 1 and Table 2. The total number of variables is 6, but only 2 variables ($x_{22}, x_{23}$) stay independent (free), which provides us an opportunity to clearly demonstrate the solutions of direct and reverse problems at the two-dimensional subspace.

Table 1
Transport Table (2 × 3)

| $c_{11}$ | $c_{12}$ | $c_{13}$ | $a_1$ |
|---|---|---|---|
| $c_{21}$ | $c_{22}$ | $c_{23}$ | $a_2$ |
| $b_1$ | $b_2$ | $b_3$ | |

Table 2
Variable Problems (2 × 3)

| $x_{11}$ | $x_{12}$ | $x_{13}$ |
|---|---|---|
| $x_{21}$ | $x_{22}$ | $x_{23}$ |

Let us express basic variables of the first column and the first line $x_{i1}, x_{1j}, (i = 1,2; j = 2,3)$ using the rest (free) variables $x_{ij}, (i = 1,2; j = 2,3)$:

$$\begin{cases} x_{11} = a_1 - b_2 - b_3 + x_{22} + x_{23} \\ x_{21} = a_2 - x_{22} - x_{23} \\ x_{12} = b_2 - x_{22} \\ x_{13} = b_3 - x_{23} \end{cases} \quad (13)$$

As all variables should be non-negative, this property should be fulfilled for the basic variables. Thus in (13) both left and right parts should be non-negative. Let us write inequality-delimitations using a standard way (accepted for LPP with a maximum). For this we should multiply parts of both inequalities by (-1):

$$\begin{cases} b_2 + b_3 - a_1 - x_{22} - x_{23} \leq 0 \\ x_{22} + x_{23} - a_2 \leq 0 \\ x_{22} - b_2 \leq 0 \\ x_{23} - b_3 \leq 0 \end{cases} \quad (14)$$

Moreover, here the non-negativity conditions should also apply: $x_{22} \geq 0; \; x_{23} \geq 0$.

Let us put basic variables into the source objective function for expressing it through two free variables. At that, for the purposes of changing the TP optimization operator from *min* to *max* (for LPP) let us change the sign of the objective function by multiplying OF of TP by (-1). In this case, within the new (free) coordinates ($x_{22}$ and $x_{23}$) the complete form of LPP's OF shall look like:

$$L(\bar{x}) = -(c_{11}(a_1 - b_2 - b_3) + c_{12}b_2 + c_{13}b_3 \\ + c_{21}a_2) \\ + (-c_{11} + c_{12} + c_{21} - c_{22})x_{22} \\ + (-c_{11} + c_{13} + c_{21} - c_{23})x_{23} \\ \to \max_{x_{ij}}. \quad (15)$$

After removing the constant component that does not influence the solution, we shall obtain a working variant of the OF:

$$L(\bar{x}) = (-c_{11} + c_{12} + c_{21} - c_{22})x_{22}$$
$$+ (-c_{11} + c_{13} + c_{21} - c_{23})x_{23}$$
$$\to \max_{x_{ij}}. \quad (16)$$

In the end, maximum LPP possessing 2 variables, which was received from the TP ($2 \times 3$), shall look like below.

Let us simplify the combination of OF coefficients by making a replacement:
$$\tilde{c}_{22} = -c_{11} + c_{12} + c_{21} - c_{22};$$
$$\tilde{c}_{23} = -c_{11} + c_{13} + c_{21} - c_{23}.$$

In this case the OF (16) shall look like:
$$L(\bar{x}) = \tilde{c}_{22}x_{22} + \tilde{c}_{23}x_{23} \to \max_{x_{ij}}, \quad (17)$$

After transforming the inequality-delimitations into the standard form, we shall receive:
$$\begin{cases} -x_{22} - x_{23} \le a_1 - b_2 - b_3 \\ x_{22} + x_{23} \le a_2 \\ x_{22} \le b_2 \\ x_{23} \le b_3 \\ -x_{22} \le 0 \\ -x_{23} \le 0 \end{cases} \quad (18)$$

Modelling Source Data

Quite often, simulation modelling is the only possible way to research the properties of the economic object management algorithms [6]. Below we shall provide the source data and the simulation modelling results pertaining to the appearing planning situations. We will choose the best transportation plan (supposedly chosen by the decision taker) and will monitor (using the equivalent coefficients) the table of expenditures, which, in new situations, can be used for making a new plan without participation of the decision taker.

Table 3 contains expenditures expressed in absolute units (e.g. in Rubles). Right table column contains one of the variants of the supply (stocks) vector value $\bar{a} = [a_1 \quad a_2]^T$, while the bottom line contains the variant of the demand vector value $\bar{b} = [b_1 \quad b_2 \quad b_3]^T$. When the line and the column intersect we can see the supply and demand balance value.

Table 3
Modelling Data

| 10 | 2  | 20 | 10 |
|----|----|----|----|
| 12 | 7  | 9  | 25 |
| 5  | 15 | 15 | 35 |

As it is shown above, the TP can be transformed into LPP, therefore they are equivalent and can be transformed into each other using a one-one principle.

As is known, left part delimitation coefficients and the OF coefficients represent vector coordinates that are normal in relation to the corresponding lines (hyperplanes). Generally speaking, lengths of these normal vectors can be random, being defined by the values of the left parts. However, as is known, the inequality (or OF) will not change should both its parts be divided with the same positive number. Should the vector's original length (specific for each delimitation and the OF) be such a number, then all lines (hyperplanes) of the delimitations and the OF become comparable, i.e. they correspond to the unit length normal vectors (ULNV), thus becoming the normalized ones. It is necessary to mention that when solving direct LPP (DLPP) it does not matter whether the delimitations and/or OF are normalized. All parameters can be in their original form, or some of the parameters can be normalized while others are not. Making non-normalized delimitations and OF normalized is important for solving reverse LPP (RLPP) [4].

Normalized coefficients $\tilde{c}_{22}$ and $\tilde{c}_{23}$ shall represent the coordinates of the vector that is normal in relation to the line (hyperplane) of the objective function (17), i.e. UNLV $\bar{e} = [e_1 \quad e_2]^T$.

The formulae for calculating the UNLV coordinates are:
$$e_1 = \frac{\tilde{c}_{22}}{\sqrt{\tilde{c}_{22}^2 + \tilde{c}_{23}^2}}; \quad e_2 = \frac{\tilde{c}_{23}}{\sqrt{\tilde{c}_{22}^2 + \tilde{c}_{23}^2}}. \quad (19)$$

They represent the transportation expenditures, i.e. should the expenditure table of the example under review contains free variables, the coefficients shall look like: $\tilde{c}_{22} = -0.225$ and $\tilde{c}_{23} = 0.974$.

When TP contains similar situations of choice, it is usually implied that transportation costs stay the same, therefore their normalized images also stay unchanged from one stage to another. It is possible to suggest that when solving RLPP, the estimate vector $\bar{\hat{e}} = [\hat{e}_1 \quad \hat{e}_2]^T$ should result in its actual value, which is $\bar{e} = [e_1 \quad e_2]^T$ vector. Therefore, as soon as the estimates are close enough to the true values that correspond to the ones of the decision taker, we can use these estimates as adequate approximation of the decision taker's criterion preferences (for the purposes of solving the direct problem).

Normalized OF of direct LPP shall look like:
$$L(\bar{x}) = -0.225x_{22} + 0.974x_{23} \to \max_{x_{ij}}. \quad (20)$$

For the purposes of further representation and analysis let us write these delimitations as a coefficient table of left and rights parts (Table 4). Let us also add there the coefficients of the normalized OF.

Table 4

Left and Right Delimitation Parts of LPP That Is Equivalent to the Original TP

| Delimitation Number | Variables | | Condition | Right Parts |
|---|---|---|---|---|
| | $x_{22}$ | $x_{23}$ | | |
| 1 | -1 | -1 | $\leq$ | $a_1 - b_2 - b_3$ |
| 2 | 1 | 1 | $\leq$ | $a_2$ |
| 3 | 1 | 0 | $\leq$ | $b_2$ |
| 4 | 0 | 1 | $\leq$ | $b_3$ |
| 5 | -1 | 0 | $\leq$ | 0 |
| 6 | 0 | -1 | $\leq$ | 0 |
| OF | -0.225 | 0.974 | *max* | |

The reverse transport problem is solved on the basis of several stages (steps) of the research. Every new research is the situation (DMS) that consists of recurrent values (vectors) of supply and demand as well as the decision that is made by the decision taker in this situation.

Coefficients of the left-part delimitations stay the same at each stage of the research (see Table 4) with only right parts changing as they represent values of supply and demand that appear at a stage. It is necessary to mention that only 4 out of 5 elements of supply and demand vectors take part in the right parts ($b_1$ does not participate). This is explained by the fact that these elements follow the equilibrium criterion, that's why we are using only one degree of freedom. Moreover, last two inequalities contained in Table 4 or in delimitations (18) stay unchanged for all observations as they represent the non-negativity property of the desired solution or it may mean that the problem's tolerance region (TR) always lies within the first quadrant.

Table 5

Observation Sample (DMS)

| Observation Step | Supply | | Demand | | | Equilibrium |
|---|---|---|---|---|---|---|
| | $a_1$ | $a_2$ | $b_1$ | $b_2$ | $b_3$ | |
| 1 | 10 | 25 | 5 | 15 | 15 | 35 |
| 2 | 13 | 52 | 26 | 19 | 20 | 65 |
| 3 | 71 | 79 | 17 | 87 | 46 | 150 |
| 4 | 2 | 29 | 12 | 13 | 6 | 31 |
| 5 | 5 | 4 | 2 | 5 | 2 | 9 |
| 6 | 65 | 70 | 56 | 43 | 36 | 135 |
| 7 | 107 | 23 | 55 | 19 | 56 | 130 |
| 8 | 96 | 6 | 24 | 5 | 73 | 102 |
| 9 | 32 | 54 | 27 | 54 | 5 | 86 |
| 10 | 31 | 79 | 32 | 47 | 31 | 110 |
| 11 | 92 | 4 | 25 | 41 | 30 | 96 |
| 12 | 44 | 50 | 47 | 45 | 2 | 94 |
| 13 | 24 | 74 | 9 | 36 | 53 | 98 |
| 14 | 64 | 81 | 83 | 56 | 6 | 145 |
| 15 | 97 | 22 | 35 | 54 | 30 | 119 |
| 16 | 14 | 6 | 9 | 8 | 3 | 20 |
| 17 | 90 | 4 | 12 | 51 | 31 | 94 |
| 18 | 27 | 56 | 45 | 13 | 25 | 83 |
| 19 | 78 | 66 | 52 | 48 | 44 | 144 |
| 20 | 75 | 99 | 65 | 52 | 57 | 174 |
| 21 | 12 | 1 | 6 | 4 | 3 | 13 |
| 22 | 31 | 69 | 24 | 44 | 32 | 100 |
| 23 | 64 | 39 | 38 | 34 | 31 | 103 |
| 24 | 83 | 36 | 28 | 51 | 40 | 119 |
| 25 | 15 | 12 | 16 | 1 | 10 | 27 |
| Polygon | 5 | 3 | 4 | 2 | 2 | 8 |

Given that the TR's current spectrum vectors play an important role in the solution of reverse TP, we shall briefly explain the meaning of the spectrum (according to [4]). TR spectrum is a cluster of vectors (UNLV) where each of them is orthogonal to one (its own) hyperplane that is included into the number of the hyperplanes that create the TR.

Therefore, the TP being transformed into LLP is a problem with the fixed (discrete) spectrum. Here TP is similar to the production problems [4]. Still, there are some differences: coefficients of the left and right parts of the delimitations as well as the OF coefficients are not always positive. These differences result in the fact that the number of active delimitations (those that create the TR) does not always include two latter ones (see (18)), which means that TR can be "hanging" in the first quadrant without touching the coordinate axis. But the UNLV of OF is turned at any other side. Besides, all this diversity of TR and OF's UNLV is defined only by the values of $\bar{a} = [a_1 \ a_2]^T$ and $\bar{b} = [b_1 \ b_2 \ b_3]^T$ vector elements. Here we should mention that the spectrum has its own characteristics in the TP. Thus, the discrete spectrum for the given dimensionality $(m \times n)$ can be random in the production problems (it can vary from one problem to another, being defined only by the matrix of the left-part delimitations). But in the TP with the concrete dimensionality (for example, the one that is reviewed here: $2 \times 3$) the spectrum is defined *only by dimensionality*, being independent from the coefficient values of the expenditure table. Only the OF of LPP that is built according to TP, depends on them.

Table 6

Decisions Made by Decision Taker (Simulation) in relation to the Observation Sample

| Observation Step | Solution | | | | | | OF | Active Delimitations | |
|---|---|---|---|---|---|---|---|---|---|
| | $x_{11}$ | $x_{12}$ | $x_{13}$ | $x_{21}$ | $x_{22}$ | $x_{23}$ | L norm_2 | Del. 1 | Del. 2 |
| 1 | 0 | 10 | 0 | 5 | 5 | 15 | 8.963 | 1 | 4 |
| 2 | 0 | 13 | 0 | 26 | 6 | 20 | 20.077 | 1 | 4 |
| 3 | 0 | 71 | 0 | 17 | 16 | 46 | 31.263 | 1 | 4 |
| 4 | 0 | 2 | 0 | 12 | 11 | 6 | 10.003 | 1 | 4 |
| 5 | 0 | 5 | 0 | 2 | 0 | 2 | 1.864 | 1 | 4 |
| 6 | 22 | 43 | 0 | 34 | 0 | 36 | 37.214 | 4 | 5 |
| 7 | 55 | 19 | 33 | 0 | 0 | 23 | 52.164 | 2 | 5 |
| 8 | 24 | 5 | 67 | 0 | 0 | 6 | 58.940 | 2 | 5 |
| 9 | 0 | 32 | 0 | 27 | 22 | 5 | 21.045 | 1 | 4 |
| 10 | 0 | 31 | 0 | 32 | 16 | 31 | 30.008 | 1 | 4 |
| 11 | 25 | 41 | 26 | 0 | 0 | 4 | 31.836 | 2 | 5 |
| 12 | 0 | 44 | 0 | 47 | 1 | 2 | 24.272 | 1 | 4 |
| 13 | 0 | 24 | 0 | 9 | 12 | 53 | 25.706 | 1 | 4 |
| 14 | 8 | 56 | 0 | 75 | 0 | 6 | 41.086 | 4 | 5 |
| 15 | 35 | 54 | 8 | 0 | 0 | 22 | 29.255 | 2 | 5 |
| 16 | 6 | 8 | 0 | 3 | 0 | 3 | 4.983 | 1 | 4 |
| 17 | 12 | 51 | 27 | 0 | 0 | 4 | 28.610 | 2 | 5 |
| 18 | 14 | 13 | 0 | 31 | 0 | 25 | 27.355 | 4 | 5 |
| 19 | 30 | 48 | 0 | 22 | 0 | 44 | 37.859 | 4 | 5 |
| 20 | 23 | 52 | 0 | 42 | 0 | 57 | 48.436 | 4 | 5 |
| 21 | 6 | 4 | 2 | 0 | 0 | 1 | 4.195 | 2 | 5 |
| 22 | 0 | 31 | 0 | 24 | 13 | 32 | 26.136 | 1 | 4 |
| 23 | 30 | 34 | 0 | 8 | 0 | 31 | 26.638 | 4 | 5 |
| 24 | 28 | 51 | 4 | 0 | 0 | 36 | 28.179 | 2 | 5 |
| 25 | 14 | 1 | 0 | 2 | 0 | 10 | 9.178 | 4 | 5 |
| Polygon | 3 | 2 | 0 | 1 | 0 | 2 | | 4 | 5 |

Observations: DMS and Decisions Made by Decision Taker

In each observation, DMS are represented by the values of two vectors, which are $\bar{a} = [a_1 \ a_2]^T$ and $\bar{b} = [b_1 \ b_2 \ b_3]^T$. Let the observation sample for the simulation experiment consist of 25 situations (DMS) where it is necessary to build the transportation plan. In Table 5 we can see the data that is obtained with the help of the random number generator (Data Analysis add-in within MS Excel environment). We generated the numbers within [1:100] interval per each column of supply and demand (excluding the last ones - $a_2$ and $b_3$), after which, in order to provide the equilibrium, we computed the rest 2 columns and/or corrected original random numbers should it be necessary.

Let us decide that the decision taker (i.e. the OF (20) that simulates his/her choice) chose values of variables $x_{ij}$ per each DMS. These step-by-step decisions are shown in Table 6. There (for the purposes of further analysis) we also provide OF values (in the normalized form - until (20)), giving as well the numbers of two delimitations (numbering is made according to Table 4) that create the extreme point, chosen by the decision taker as the optimal one. These delimitations are called "active delimitations" because they participate in forming the optimal point, i.e. the solution for the given DMS.

In the last line of Table 5 and Table 6 we provide a polygon that corresponds to the specific DMS that was built according to the problem's spectrum (left parts of delimitations). Further we shall use this DMS for checking quality of the objective function.

In spite of a seemingly big and possible variety of TR variants, TP-caused LLP, unlike other types of linear programming models (e.g. production-type), possess very special characteristics. Let us consider special characteristics pertaining to this type of problems.

*Possible TR-Configurations for the Problem under Review.*

Pictures 1 to 20 show all possible TR configurations of LPP for TP (2 × 3). The figures represent numbers of delimitations (numbering is made according to Table 4). Last five DMS (Pic. 16 - Pic. 20) are different from the rest because of the special value combination of the transport table, i.e. the 1st delimitation is placed beyond the first quadrant, thus not participating in the TR creation. Generally speaking, there can be TRs that generate into intervals, i.e. when some pair of parallel delimitations coincides, e.g. 1-2, 3-5 or 4-6. However, such situations are very rare. Therefore we shall consider them generated and will exclude them from the further consideration. We shall also not consider the situations where the extreme point is created by three and not by two lines. Such cases are rather rare and should they happen, it is always possible to choose the pair of lines that are directly adjacent to TR.

Peculiarity Analysis of LPP Built on the Basis of TP

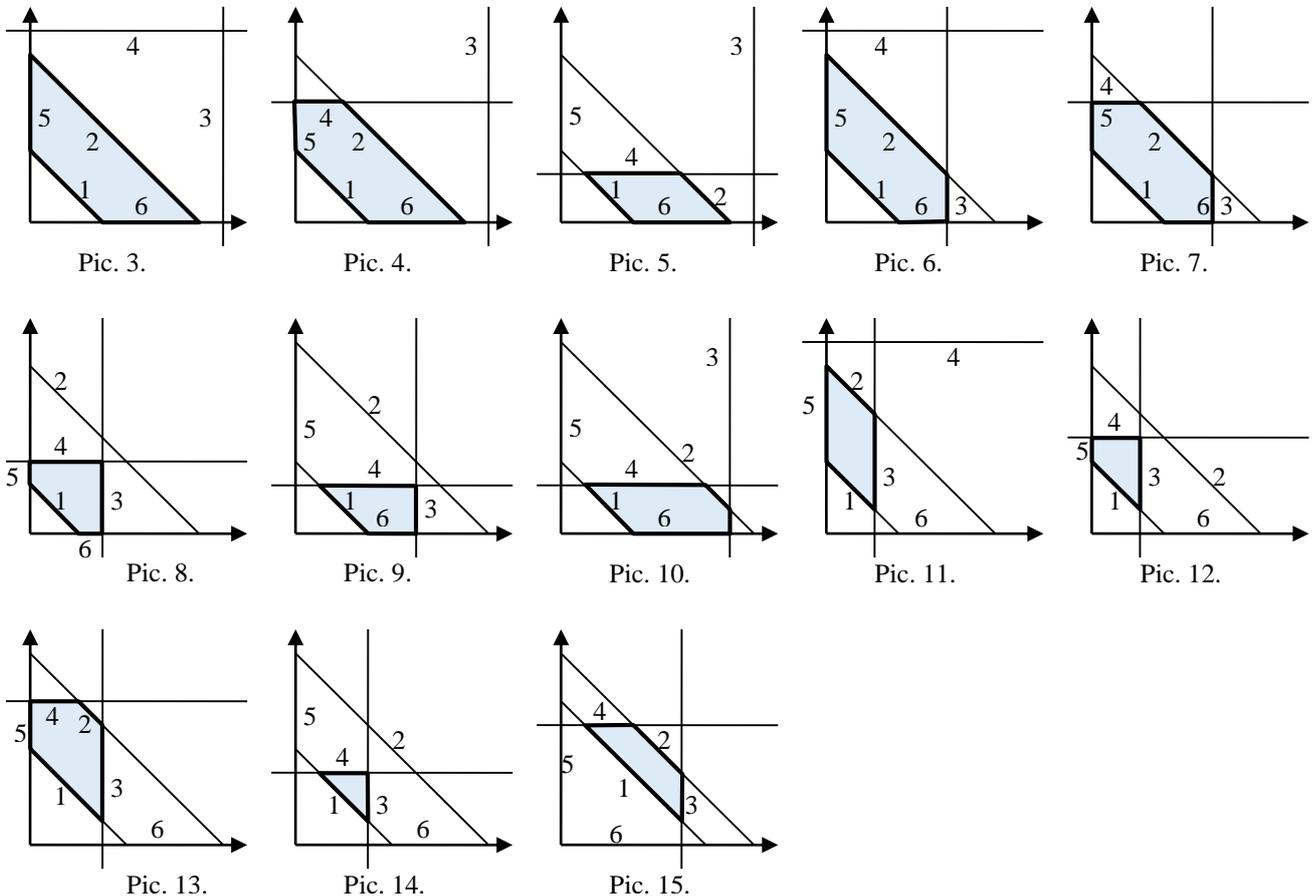

Pic. 3. Pic. 4. Pic. 5. Pic. 6. Pic. 7.

Pic. 8. Pic. 9. Pic. 10. Pic. 11. Pic. 12.

Pic. 13. Pic. 14. Pic. 15.

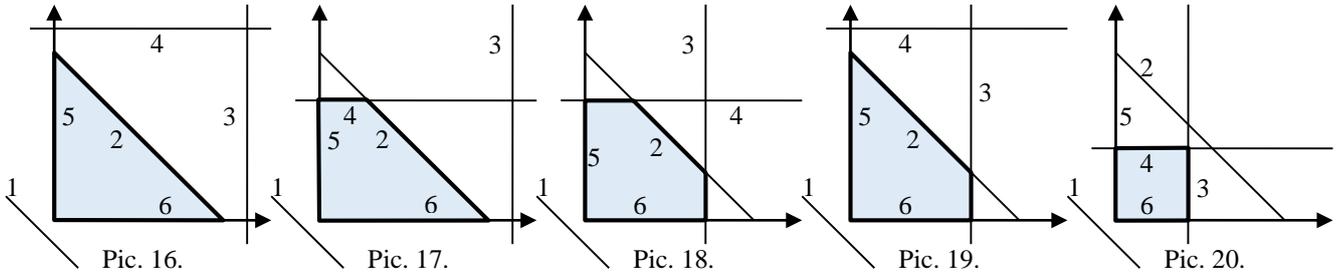

| Pic. 16. | Pic. 17. | Pic. 18. | Pic. 19. | Pic. 20. |

*Polygon.*

The polygon is [4] such a TR (see Pic. 21), which possesses many important and special properties that allow it to use a corresponding DMS as a control situation to check the quality of model's settings in relation to the observations and other types of research.

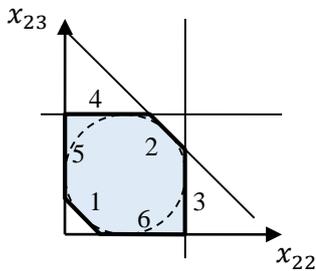

Pic. 21. TR-Polygon

The polygon includes the following important properties:
- All delimitations of the polygon are active, i.e. they participate in the TR border formation.
- All TR alternatives are maximally informative. In the case of 2-dimensional situation it stands for the maximum possible obtuse angles at the extreme TR points.
- The alternatives are evenly (maximally) contrast, i.e. ideally the paired distances between the alternatives (at the TR border) are the same. Such variant is not always easy to perform technically, but we should strive to it. Quite often the compromise is the polygon where separate lines (hyperplanes) of delimitations are the tangents to some circumference (hypersphere). This variant is shown on Pic. 21.

The polygon is used when researching the model parameter setting process, playing the role of a "litmus test paper". It checks whether the decision made on the basis of the set-up model corresponds to the decision made by the decision taker (or his/her simulation).

Peculiarity of TP lies in the fact that type of the polygon does not depend on data, depending only on the problem's dimensionality. Therefore, for all values of supply and demand vectors the polygon shall look like Pic. 21.

*Spectra of the Problem (Delimitations, OF) and Observations.*

On Picture 22 we can see the problem's spectrum, i.e. the UNLV population of six lines of delimitations and one line of OF level.

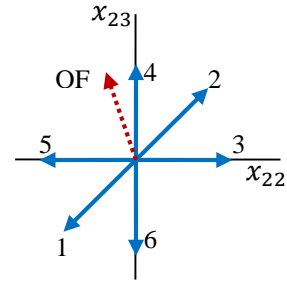

Pic. 22. TR-Polygon

It is necessary to mention that TP's UNLV of OF can be directed anywhere (unlike production-type LPP where UNLV of OF can lie only within the first quadrant).

*Possible Types of Decisions (for Random OF). Pairs of Spectral Vectors.*

For the TR that corresponds to the situation in place (DMS), the decision taker (or OF that simulates him/her) chooses one of the extreme points as a solution. One of UNLV pairs corresponds to the pair of delimitation lines that form the chosen extreme point (see Pic. 22). If to consider the potentially possible variants of UNLV pairs that can participate in the formation of extreme points, we shall see 6 two-figure combinations. Here we do not include three pairs of UNLV that are parallel to each other (1-2, 3-5 and 4-6). This multitude consists of 12 variants: 1-3, 1-4, 1-5, 1-6, 2-3, 2-4, 2-5, 2-6, 3-4, 3-6, 4-5, 5-6. The decision taker chooses one of the TR's extreme points formed by the corresponding UNLV pair as an optimal one. Variants of these pairs (arrows in bold) are shown in Table 7. Here we can also see sum vectors (double arrows) for each UNLV pair. All 12 pairs are divided into three groups that are distinguished by the angle between the paired vectors and consequently, by the length of the sum vector. Looking ahead, we can notice that the length of the sum vector reflects the informativeness of the solution, i.e. of that extreme TR point, to which the given UNLV pair corresponds. The sum vector length is used as the weights of the corresponding observations when solving the reverse problem. Thus, all possible observations can belong to one of three groups: to the 1st group - the least informative, to the 3rd group - the most informative and to the 2nd group - medium informative.

Table 7

UNLV Pairs

| Groups | UNLV Pairs in Groups | | | |
|---|---|---|---|---|
| 1 | 1-3 | 1-4 | 2-5 | 2-6 |
| 2 | 3-4 | 3-6 | 4-5 | 5-6 |
| 3 | 1-5 | 1-6 | 2-3 | 2-4 |

Apart from DMS, the observation includes the decision that was made and which corresponds to one of TR's extreme points, i.e. to the sum vector, shown as a double arrow at the pictures of Table 7. Eventually, the result of each observation is the only vector (*observation vector*), which is used in the solution algorithm of the reverse problem for estimating the decision taker's OF vector. Only direction of the desired decision taker's OF vector is important, its length is not of any importance. As far as observation vectors are concerned, their direction and length are of a big interest because the length shows the informative value of the given observation, i.e. its contribution into the evaluation process of OF performed by the decision taker. On Picture 23 we can see all observation vectors (sums of vector pairs) that are available for the TP example under review. Thus, in each observation the DMS is represented as some population (from three to six) of the active delimitation vectors. The decision that was made by the decision taker is shown in the following manner: one of these vectors is marked by a dot like on Pic. 23 where out of 25 observations of the example under review we chose only three as the optimal ones: 1-4, 2-5, 4-5. It is necessary to mention though that pairs 1-4 and 2-5 have minimum weights (lengths), while pair 4-5 has an average weight. On Pic. 24 we can see UNLVs both of observations and the decision taker simulated OF.

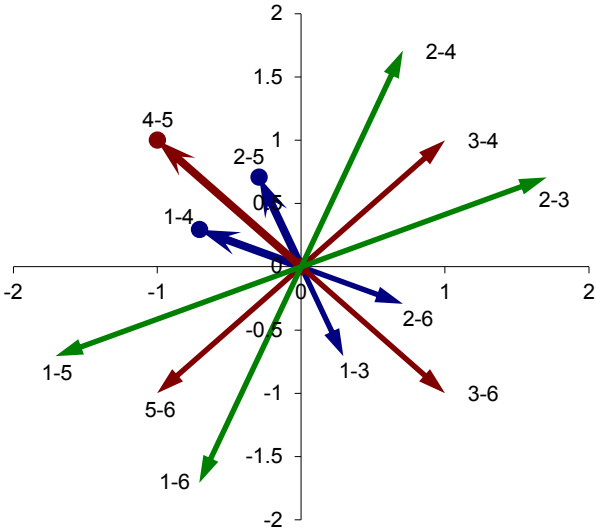

Pic. 23. Variants of Observational Vectors

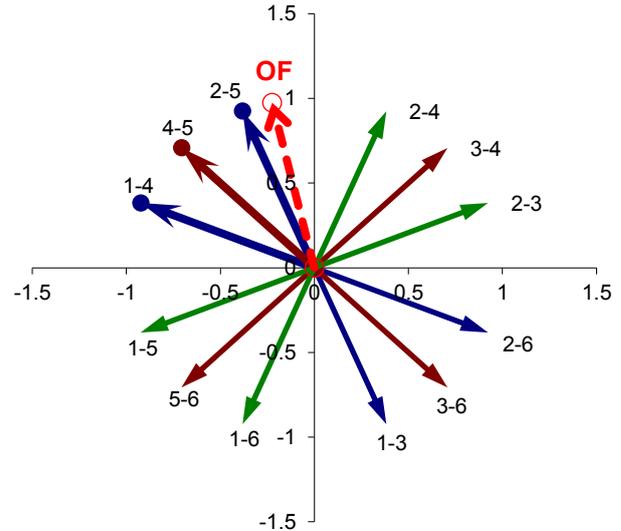

Pic. 24. UNLV of Observations and OF

*Comparative Characteristics of the Observation Variants.*

Different TR configurations, shown on Pictures 3-20 differ from each other on the basis of some properties' values (number of alternatives, their informativeness etc.).

In Table 8 we can see their combined characteristics.

Table 8

Properties of Observational Variants

| No. | Type of TR | Active Delimitations | | Informativeness Indices | | | | Groups of One-Type TR |
|---|---|---|---|---|---|---|---|---|
| | | Quantity | Numbers | Alternative's Ranks | General Rank | Average Rank | Average Weight | |
| 1 | Pic. 3. | 4 | 1,2,5,6 | 1, 1, 3, 3 | 8 | 2 | 0.347 | 5 |
| 2 | Pic. 4. | 5 | 1,2,4,5,6 | 1, 2, 3, 3, 3 | 12 | 2.4 | 0.444 | 7 |
| 3 | Pic. 5. | 4 | 1,2,4,6 | 1, 1, 3, 3 | 8 | 2 | 0.347 | 4 |
| 4 | Pic. 6. | 5 | 1,2,3,5,6 | 1, 2, 3, 3, 3 | 12 | 2.4 | 0.444 | 7 |
| **5** | **Pic. 7.** | **6** | **1,2,3,4,5,6** | **2, 2, 3, 3, 3, 3** | **16** | **2.7** | **0.509** | **9** |
| 6 | Pic. 8. | 5 | 1,3,4,5,6 | 2, 2, 2, 3, 3 | 12 | 2.4 | 0.423 | 8 |
| 7 | Pic. 9. | 4 | 1,3,4,6 | 1, 2, 2, 3 | 8 | 2 | 0.320 | 3 |
| 8 | Pic. 10. | 5 | 1,2,3,4,6 | 1, 2, 3, 3, 3 | 12 | 2.4 | 0.444 | 6 |
| 9 | Pic. 11. | 4 | 1,2,3,5 | 1, 1, 3, 3 | 6 | 2 | 0.347 | 4 |
| 10 | Pic. 12. | 4 | 1,3,4,5 | 1, 2, 2, 3 | 8 | 2 | 0.320 | 3 |
| 11 | Pic. 13. | 5 | 1,2,3,4,5 | 1, 2, 3, 3, 3 | 12 | 2.4 | 0.444 | 6 |
| 12 | Pic. 14. | 3 | 1,3,4 | 1, 1, 2 | 4 | 1.3 | 0.148 | 1 |
| 13 | Pic. 15. | 4 | 1,2,3,4 | 1, 1, 3, 3 | 8 | 2 | 0.347 | 5 |
| 14 | Pic. 16. | 3 | 2,5,6 | 1, 1, 2 | 4 | 1.3 | 0.148 | 1 |
| 15 | Pic. 17. | 4 | 2,4,5,6 | 1, 2, 2, 3 | 8 | 2 | 0.320 | 3 |
| 16 | Pic. 18. | 5 | 2,3,4,5,6 | 2, 2, 2, 3, 3 | 12 | 2.4 | 0.423 | 8 |
| 17 | Pic. 19. | 4 | 2,3,5,6 | 1, 2, 2, 3 | 8 | 2 | 0.320 | 3 |
| 18 | Pic. 20. | 4 | 3,4,5,6 | 2, 2, 2, 2 | 8 | 2 | 0.293 | 2 |

In Table 8 the alternative's rank is a whole number ($r=1,2,3$) that can have one value out of three: $r = 1$ with $w = 0.076$; $r = 2$ with $w = 0.293$; $r = 3$ with $w = 0.617$, where $w$ is the observational weight. General rank is the total sum of ranks of all alternatives related to this observation. Average rank is the rank that was averaged according to the multitude of alternatives related to this observation. Average weight is the averaged weight value related to the multitude of observations.

Average rank or average weight define the informative value of this observation, i.e. its contribution into the information gain about the evaluated OF made by the decision taker who can actually perform this observation. It is clear that the TR, whose configuration is similar to the polygon (Pic. 21) possesses a larger informative value (Pic. 7).

In Table 8 we can see the groups of one-type TRs, inside of which the situation changes probably on the basis of the region's turn. These groups are numbered according to the increase of their average weight (or average rank).

It is necessary to pay attention to the 5[th] observation. All its characteristics are outstanding: it possesses maximal (in comparison with other observational variants) average weight and average rank together with the maximum number of active delimitations, i.e. the delimitations that form the TR. In this case *all* existing delimitations take place in the process. *Polygon* also looks like this delimitation.

Solution of Reverse Problem (Restoration of OF Parameters on the basis of Observations)

The main computational formula of a *single-point step-by-step algorithm* used for the estimation of model's parameters within the observations looks like [4]:

$$\hat{c}_k^i = \frac{1}{\sqrt{\left(\sum_{j=1}^k \beta_j e_j^1\right)^2 + \left(\sum_{j=1}^k \beta_j e_j^2\right)^2}} \sum_{j=1}^k \beta_j e_j^i, \quad (21)$$

where $i = 1; 2$ is the coordinate number; $\beta_j$ is the weight coefficient of *j*-th observation.

There is no discounting in this algorithm, its role is played indirectly by the coordinate normalization of the estimate vector (reduction to a single length) where the numerator uses an accumulated coordinate. Whereas the sum gradually increases within the accumulated coordinates, the relative contribution of each new observation shall decrease in proportion to this accumulation.

Should we introduce a sliding summing-up interval (e.g. using a *K* length), the summing-up limits within the sums of formula (21) shall look like: $\sum_{j=k-K+1}^{k} \cdots$ for $k > K$.

Table 9

Decisions Made by Decision Taker in relation to the Observation Sample

| Observation Step | Delimitation Pair | UNLV 1 | | UNLV 2 | | UNLV of Observation | | Observational Weight | $\sum_{j=1}^{k} \beta_j e_j^i$ | | $\hat{c}_k^i$ | |
|---|---|---|---|---|---|---|---|---|---|---|---|---|
| | | $i=1$ | $i=2$ | $i=1$ | $i=2$ | $i=1$ | $i=2$ | | $i=1$ | $i=2$ | $i=1$ | $i=2$ |
| 1 | 1-4 | -0.707 | -0.707 | 0 | 1 | -0.924 | 0.383 | 0.076 | -0.070 | 0.029 | **-0.924** | **0.383** |
| 2 | 1-4 | -0.707 | -0.707 | 0 | 1 | -0.924 | 0.383 | 0.076 | -0.140 | 0.058 | **-0.924** | **0.383** |
| 3 | 1-4 | -0.707 | -0.707 | 0 | 1 | -0.924 | 0.383 | 0.076 | -0.211 | 0.087 | **-0.924** | **0.383** |
| 4 | 1-4 | -0.707 | -0.707 | 0 | 1 | -0.924 | 0.383 | 0.076 | -0.281 | 0.117 | **-0.924** | **0.383** |
| 5 | 1-4 | -0.707 | -0.707 | 0 | 1 | -0.924 | 0.383 | 0.076 | -0.352 | 0.146 | **-0.924** | **0.383** |
| 6 | 4-5 | 0 | 1 | -1 | 0 | -0.707 | 0.707 | 0.293 | -0.559 | 0.353 | **-0.846** | **0.534** |
| 7 | 2-5 | 0.707 | 0.707 | -1 | 0 | -0.383 | 0.924 | 0.076 | -0.588 | 0.423 | **-0.812** | **0.584** |
| 8 | 2-5 | 0.707 | 0.707 | -1 | 0 | -0.383 | 0.924 | 0.076 | -0.617 | 0.493 | **-0.781** | **0.625** |
| 9 | 1-4 | -0.707 | -0.707 | 0 | 1 | -0.924 | 0.383 | 0.076 | -0.687 | 0.523 | **-0.796** | **0.605** |
| 10 | 1-4 | -0.707 | -0.707 | 0 | 1 | -0.924 | 0.383 | 0.076 | -0.758 | 0.552 | **-0.808** | **0.589** |
| 11 | 2-5 | 0.707 | 0.707 | -1 | 0 | -0.383 | 0.924 | 0.076 | -0.787 | 0.622 | **-0.784** | **0.620** |
| 12 | 1-4 | -0.707 | -0.707 | 0 | 1 | -0.924 | 0.383 | 0.076 | -0.857 | 0.651 | **-0.796** | **0.605** |
| 13 | 1-4 | -0.707 | -0.707 | 0 | 1 | -0.924 | 0.383 | 0.076 | -0.927 | 0.680 | **-0.806** | **0.591** |
| 14 | 4-5 | 0 | 1 | -1 | 0 | -0.707 | 0.707 | 0.293 | -1.135 | 0.887 | **-0.788** | **0.616** |
| 15 | 2-5 | 0.707 | 0.707 | -1 | 0 | -0.383 | 0.924 | 0.076 | -1.164 | 0.958 | **-0.772** | **0.636** |
| 16 | 1-4 | -0.707 | -0.707 | 0 | 1 | -0.924 | 0.383 | 0.076 | -1.234 | 0.987 | **-0.781** | **0.625** |
| 17 | 2-5 | 0.707 | 0.707 | -1 | 0 | -0.383 | 0.924 | 0.076 | -1.263 | 1.057 | **-0.767** | **0.642** |
| 18 | 4-5 | 0 | 1 | -1 | 0 | -0.707 | 0.707 | 0.293 | -1.470 | 1.264 | **-0.758** | **0.652** |
| 19 | 4-5 | 0 | 1 | -1 | 0 | -0.707 | 0.707 | 0.293 | -1.677 | 1.471 | **-0.752** | **0.659** |
| 20 | 4-5 | 0 | 1 | -1 | 0 | -0.707 | 0.707 | 0.293 | -1.884 | 1.678 | **-0.747** | **0.665** |
| 21 | 2-5 | 0.707 | 0.707 | -1 | 0 | -0.383 | 0.924 | 0.076 | -1.914 | 1.749 | **-0.738** | **0.675** |
| 22 | 1-4 | -0.707 | -0.707 | 0 | 1 | -0.924 | 0.383 | 0.076 | -1.984 | 1.778 | **-0.745** | **0.667** |
| 23 | 4-5 | 0 | 1 | -1 | 0 | -0.707 | 0.707 | 0.293 | -2.191 | 1.985 | **-0.741** | **0.671** |
| 24 | 2-5 | 0.707 | 0.707 | -1 | 0 | -0.383 | 0.924 | 0.076 | -2.220 | 2.055 | **-0.734** | **0.679** |
| 25 | 4-5 | 0 | 1 | -1 | 0 | -0.707 | 0.707 | 0.293 | -2.427 | 2.262 | **-0.732** | **0.682** |
| Polygon | **4-5** | 0 | 1 | -1 | 0 | **-0.707** | **0.707** | | | | **-0.707** | **0.707** |

*Evaluation Results.*

As a continuation of the above-mentioned observations (Table 5 and Table 6) let us provide in Table 9 the calculation results received according to the single-point step-by-step algorithm.

Evaluation algorithm (21) is actually an averaging procedure performed with the spectral observation vectors that are taken with the corresponding weights and which are related to the multitude of the observational steps. Considering frequencies of three spectral vectors that are observed during 25 steps, it is clear that their average value should be formed within the neighborhood of 4-5 spectral vector. The calculations, provided in Table 9, confirmed this conclusion. It is necessary to note that the estimate is getting close to the Polygon's observation vector (see Polygon line in Table 9).

Explanation of the Obtained Results

Algorithm (21) is built on the basis of averaging the weighted observation vectors. Thus, if the data was formed randomly, the appearances of any observational vector (see Pic. 23) are equally possible. As it can be seen on Pictures 23 and 24, not all directions possess equal *informativeness.* Therefore, the wide pattern of the observation spectrum leads to the fact that the final setting-up vector is displaced in relation to the actual (modelled) OF vector of the decision taker. However, the wide pattern of the observation spectrum can also play a positive role: thus, if the representative spectral line is "taken", it will provide high quality of the decisions made in the future.

The estimate convergence is shown graphically on Pic. 25.

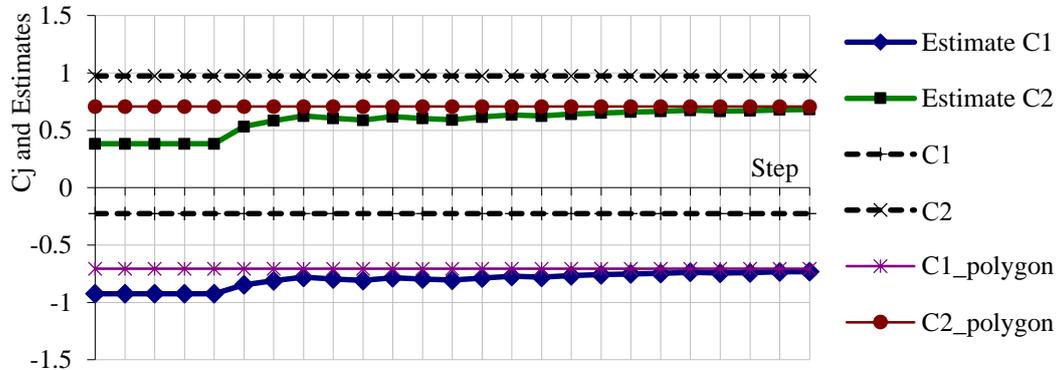

Pic. 25. Convergence of Decision Taker's OF Estimates

Here it can be seen that the estimates of OF coefficients (its UNLV) are converged to the values of the solution (its UNLV) at the Polygon and not to the actual (modelled) values. Still, according to the modelling results, we can see that within any newly appearing DNS (see Pic. 3 - Pic. 20) the solutions, which are obtained with the adjusted (estimated) OF do not lead to errors (they correspond to the solutions obtained in relation to the modelled OF). All model-type solutions that are accepted in any DMS shall correspond to the solutions, accepted by the decision taker (that simulate his/her OF). Thus, the approximation of OF, which was not estimate-effective, turned out to be a *solution-effective* one.

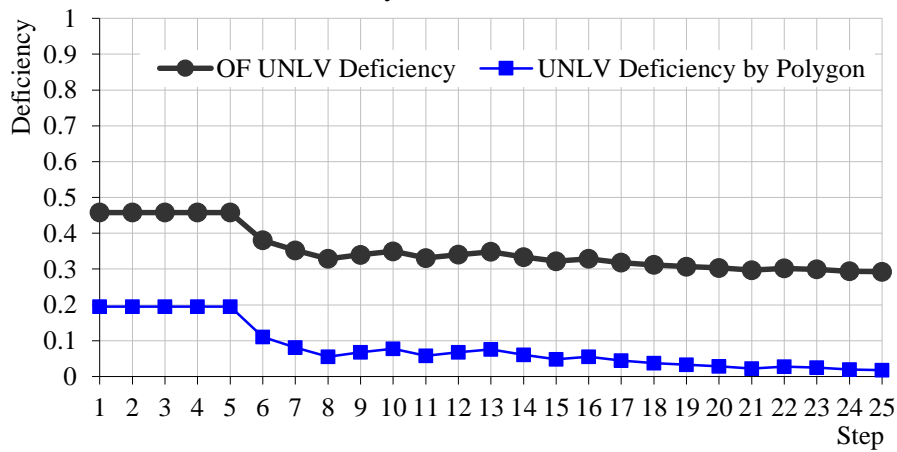

Pic. 26. Convergence of Vectors' Differences

Deficiency charts showing the estimate vector of approximating OF in relation to the actual OF of UNLV (upper one) and in relation to the closest spectral vector of the Polygon are shown on Picture 26.

Here we do not provide the solution convergence because right after the first step of settings the model's estimates turned out to be accurate enough for the solutions accepted for all other appearing DMS to completely correspond between each other according to the OF of the decision taker and according to the set-up model (its approximation).

On Adequacy Logic of the Reestablished Model

Presence of fast solution-convergence and bad estimate-convergence is explained by the fact that the decision taker's OF becomes apparent only through DMS (TR). Out of all possible DMS only the DMS-polygon is the most representative and the most informative (see Table 8) representative of the environment, where the decision taker works. External observer sees objective preferences (OF-shaped) of the decision taker through DMS, therefore the OF of the decision taker should look like one of DMS elements. Problem spectrum or polygon spectrum vectors are such DMS elements. In the process of reestablishment (estimation) of the decision taker's OF we can find the UNLV of OF (as an image of OF), approximating it with one of the polygon spectrum vectors. Thus we can talk about approximating OF of the decision taker using one of the polygon *observation vectors* (see vectors 2-4, 2-3, 3-6, 1-6, 1-5 and 4-5 on Pic. 24). Thus, OF of the decision taker, represented by the continuous UNLV (Pic. 24) and being projected at DMS is discretized by the problem's spectrum, whose full informational representative is the polygon's spectrum. This is why the search for the estimate of the decision taker's OF that approximates his/her preferences can be performed only at the discrete spectrum of the problem (polygon). This explains the fact that the estimates (UNLV of OF) converge to one of the polygon's spectrum vectors and not to the continuous and real UNLV of decision taker's OF.

It is also necessary to note that quality of approximation depends on the representativeness degree of DMS-multitude at the estimation stage, i.e. how completely it reflects the variety of all possible situations. If DMS-multitude is representative (adequate to the environment), we can talk about approximation that is adequate to any potentially possible DMS. If DMS-multitude reflects only a part of possible situations, then here we see local approximation, where we use only a part of the problem's spectrum or the polygon's spectrum (i.e. only a local spectrum is used) within the setting-up procedure and in the course of the further solution of the direct TP. In this case, the solutions obtained on the basis of the adjusted model shall be reliable only for the new DMS, which appears within the same local region of the spectrum. It is possible to say that this search will be performed "with the light" in that part of the problem's spectrum, which was already "lit up" by the previous set-up steps. If there appears a DMS that extends beyond the borders of the local one, it is necessary to test decision taker again for the knowledge of this new area, afterwards correcting the OF estimates.

**Conclusions**

1. Modelling approximation process of the decision taker preferences within the transport system using general transport table shows high speed of the solution convergence, thus providing grounds for application of similar approximations in the transportation planning systems as well as in other applications, described by the scheme of the transport problem.

2. Research of both the approximation algorithm and properties of the constructed model of the transport problem showed that solutions obtained with the help of the adjusted model can possess local effectiveness, i.e. solutions of the direct transport problem, obtained with the help of the adjusted model can be as good as the solutions, obtained by the decision taker in the same situations.

3. Stopping rules of the model's set-up process can be based on the statistic characteristics of the variations pertaining to the estimate vector of decision taker's OF such as average value and average quadratic deviation.